\pdfoutput=1

\documentclass[11pt]{article}

\usepackage[]{emnlp2021}

\usepackage{times}
\usepackage{latexsym}

\usepackage[T1]{fontenc}

\usepackage[utf8]{inputenc}

\usepackage{microtype}

\usepackage{graphicx}
\usepackage{multirow}
\usepackage{microtype}
\usepackage{multicol}
\usepackage{footmisc}
\usepackage{ctable}
\usepackage{amsmath}
\usepackage{newtxtext,newtxmath}
\usepackage[labelfont=bf]{caption}
\usepackage{subcaption}

\newcommand\blfootnote[1]{%
  \begingroup
  \renewcommand\thefootnote{}\footnote{#1}%
  \addtocounter{footnote}{-1}%
  \endgroup
}

\title{Weakly-supervised Multi-task Learning for Multimodal Affect Recognition}

\author{Wenliang Dai, Samuel Cahyawijaya, Yejin Bang, Pascale Fung \\
Center for Artificial Intelligence Research (CAiRE)\\
Department of Electronic and Computer Engineering\\
The Hong Kong University of Science and Technology, Clear Water Bay, Hong Kong\\
\texttt{\{wdaiai,scahyawijaya,yjbang\}@connect.ust.hk}} 

\begin{document}
\maketitle
\begin{abstract}
Multimodal affect recognition constitutes an important aspect for enhancing interpersonal relationships in human-computer interaction.
However, relevant data is hard to come by and notably costly to annotate, which poses a challenging barrier to build robust multimodal affect recognition systems.
Models trained on these relatively small datasets tend to overfit and the improvement gained by using complex state-of-the-art models is marginal compared to simple baselines.
Meanwhile, there are many different multimodal affect recognition datasets, though each may be small. In this paper, we propose to leverage these datasets using weakly-supervised multi-task learning to improve the generalization performance on each of them. Specifically, we explore three multimodal affect recognition tasks: 1) emotion recognition; 2) sentiment analysis; and 3) sarcasm recognition. Our experimental results show that multi-tasking can benefit all these tasks, achieving an improvement up to 2.9\% accuracy and 3.3\% F1-score. Furthermore, our method also helps to improve the stability of model performance.
In addition, our analysis suggests that weak supervision can provide a comparable contribution to strong supervision if the tasks are highly correlated.
\blfootnote{Preprint. Code will be available soon.}
\end{abstract}

\section{Introduction} \label{sec:intro}
Deep learning involving multiple modalities can be seen as a joint field of computer vision and natural language processing, which has become much more popular in recent years~\cite{Vinyals2015ShowAT,vqa2,Sanabria2018How2AL}. For human affect recognition tasks (e.g. emotion recognition, sentiment analysis, sarcasm recognition, etc.), more modalities can provide complementary and supplementary information~\citep{baltruvsaitis2018multimodal} and help to recognize affect more accurately.
Prior works mainly focus on two major directions: 1) modeling the \textit{intra}-modal dynamics (unimodal representation learning); and 2) improving the \textit{inter}-modal dynamics (cross-modal interactions and modality fusion). 
For example, various fusion methods have been proposed, ranging from the basic model-agnostic ones like Early-Fusion~\citep{morency2011towards} and Late-Fusion~\citep{zadeh2016mosi}, to more complex ones like Tensor-based Fusion~\citep{Zadeh2017TensorFN,Liu2018EfficientLM}, Attention-based Fusion~\citep{mfn,raven,mult}.

\begin{figure}[t]
    \centering
    \includegraphics[width=\linewidth]{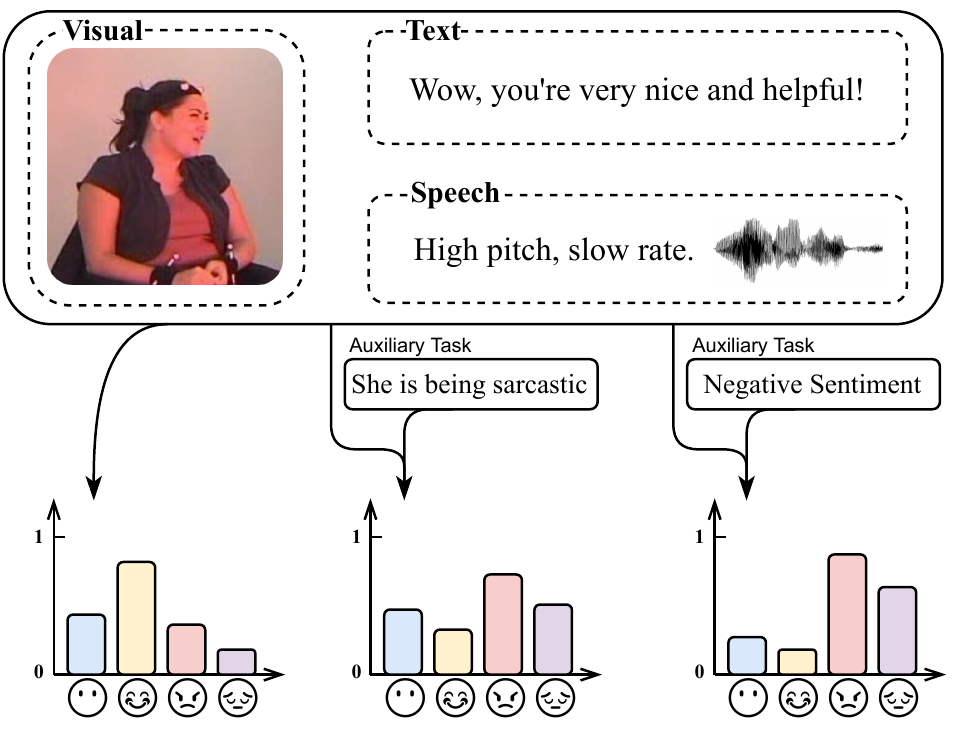}
    \caption{An example of  multi-task learning (MTL) when emotion recognition is the main task. Given the input (positive text, skeptical face and hesitating voice), it is not clear what emotion the woman has, resulting in a positive prediction due to the strong signal from the textual modality. However, if we could know she is actually being sarcastic or her sentiment is negative, the prediction will be leaned towards negative emotions. MTL is a way to eavesdrop external information by leveraging more useful supervisions. In this paper, we explore weakly-supervised MTL which enables to get more supervisions with zero extra human labor.}
    \label{fig:intro}    
\end{figure}

Despite the progress, multimodal affect recognition datasets~\citep{iemocap,zadeh2016mosi,zadeh2018multimodal,mustard} present a performance bottleneck that they are relatively small when compared with unimodal research datasets~\citep{Deng2009ImageNetAL,Lin2014MicrosoftCC,squad1} or other multimodal ones~\citep{mscoco,clevr,Sidorov2020TextCapsAD},
which poses two potential problems: 1) models are easy to overfit and cannot generalize well; and 2) instability of performance (e.g. alter a random seed for weights initialization can lead to a salient change of the model performance).
To study and validate the severity these problems, we benchmark 12 different models on two commonly used multimodal emotion recognition datasets (Table~\ref{tab:iemocap-result-full} and Appendix A). The experimental results show that recent state-of-the-art (SOTA) models do not have an obvisou advantage over simple baselines. By constructing a naive hybrid fusion mechanism, the baselines can easily surpass the SOTA models. Furthermore, to investigate the instability issue, we run each model with five different random seeds (0 - 4) on those two datasets. Averagely, the standard deviations of the weighted accuracy~\citep{wacc} and F1 are 1.8\% and 1.7\%, respectively. 
The performance improvement gained by just altering a random seed is similar to or even larger than the improvement gained by using a more advanced method from the prior work~\citep{mfn,mult,wang2019words,Dai2020ModalityTransferableEE}.

Given the fact that there are many different multimodal datasets related to affect recognition and most of them are single-task labeled, to mitigate the aforementioned problems, we propose to conduct weakly-supervised multi-task learning (MTL) to improve the generalization performance and stability of models~\citep{mtl_survey} (Figure \ref{fig:intro}). Compared to MTL with strong supervision, weakly-supervised MTL does not require the datasets to have multiple labels at the same time, which makes it much cheaper and more flexible.
It can be seen as an implicit way of achieving data augmentation without any additional human labor (more details in Section~\ref{sec:mtl}). According to our experiments, we get an improvement up to 2.9\% accuracy and 3.3\% F1-score by leveraging weakly-supervised MTL.

The contributions of this paper are summarized as follows:
\begin{itemize}
    \item We thoroughly benchmark 12 models on two widely used multimodal affect recognition datasets. Based on this, we further propose a simple but effective hybrid model-agnostic modality fusion method, which performs equally or even better than previous state-of-the-art models. 
    \item We show the effectiveness of weakly-supervised MTL on relevant multimodal affect recognition tasks. We achieve an improvement, up to 2.9\% accuracy and 3.3\% F1-score, on three multimodal affect recognition tasks. Furthermore, our results demonstrate that weak labels can bring comparable improvement as strong labels. 
\end{itemize}


\section{Related Works} \label{related_works}
\paragraph{Multimodal affect recognition.} Multimodal affect recognition has attracted increasing attention in recent years. It can be seen as a family of tasks, including multimodal emotion recognition, sentiment analysis, sarcasm recognition, etc. There are two major focuses in this research field~\citep{baltruvsaitis2018multimodal}: 1) how to better model intra-modal dynamics, i.e. improving the representation learning of a single modality; and 2) how to improve the inter-modal dynamics, i.e. the interactions cross different modalities.
Diverse methods have been proposed to improve these two parts. For example, quite a few works focus on the fusion of modalities, such as the Tensor Fusion Network~\citep{Zadeh2017TensorFN}, Memory Fusion Network~\citep{mfn}, Multimodal Adaptation Gate~\citep{rahman2020integrating}. Additionally, Multimodal Transformer~\citep{mult} was introduced to handle unaligned data, \citet{Dai2020ModalityTransferableEE} proposed to use emotional embeddings to enable zero-/few-shot learning for low-resource senarios, and \citet{Dai2021MultimodalES} introduced the sparse cross-attention to improve performance and reduce computation. Despite the remarkbale progress has been made, we find that most models suffer from the small scale of data on these tasks. For example, the Multimodal Transformer~\citep{mult} has a dimenion of only 40, meanwhile with a large dropout value around 0.3.

\paragraph{Multi-task Learning (MTL).} MTL has been widely used in numerous tasks to improve the performance of models. It can be seen as an implicit data augmentation and an eavesdropping of extra supervision to improve the generalization ability of models~\citep{mtl_survey}.
For example, in computer vision, \citet{Xu2018PADNetMG} proposed PAD-Net, which tackles depth estimation and scene parsing in a joint CNN with four intermediate auxiliary tasks. Moreover, \citet{Kokkinos2017UberNetTA} invented UberNet, which solves seven different tasks simultaneously. In natural language processing, MTL is also leveraged in various tasks, such as offensive text detection~\citep{abu-farha-magdy-2020-multitask,dai-etal-2020-kungfupanda}, summarization~\citep{yu-etal-2020-dimsum}, question answering~\citep{McCann2018TheNL}, etc.

For multimodal affect recognition tasks, not much work has been carried out to incorporate multi-tasking.
Of the few works that have considered multi-tasking, \citet{Akhtar2019MultitaskLF} proposed to tackle sentiment analysis and emotion recognition jointly. However, their method is only verified on one dataset, as it requires the dataset to have multiple human annotation at the same time. Also, \citet{Chauhan2020SentimentAE} studied the relationship between sentiment, emotion, and sarcasm by manually annotating a sarcasm dataset~\citep{mustard}, which we believe does not scale. In addition, they only labeled a few hundred of samples, which limits the persuasion of their results. Differently, our work explores weakly-supervised multi-task learning which is much cheaper and more scalable. Moreover, our experiments are done in a wider range with three different datsets.

\section{Data and Evaluation Metrics}
In this section, we first introduce three datasets used for model benchmarking and weakly-supervised multi-task learning. Then, we discuss the feature extraction algorithms to pre-process the data. Finally, we illustrate the metrics we use to evaluate models.

\subsection{Datasets}

\paragraph{IEMOCAP.} The Interactive Emotional Dyadic Motion Capture (IEMOCAP)~\citep{iemocap} is a dataset for multimodal emotion recognition, which contains 151 videos along with the corresponding transcripts and audios. In each video, two professional actors conduct dyadic dialogues in English. Although the human annotation has nine emotion categories, following the prior works~\citep{Hazarika2018ConversationalMN,raven,mult,Dai2020ModalityTransferableEE}, we take four categories: \textit{neutral}, \textit{happy}, \textit{sad}, and \textit{angry}. 

\paragraph{CMU-MOSEI.} The CMU Multimodal Opinion Sentiment and Emotion Intensity (CMU-MOSEI)~\citep{zadeh2018multimodal} is a dataset for both multimodal emotion recognition and sentiment analysis. It comprises 3,837 videos from 1,000 diverse speakers and annotated with six emotion categories: \textit{happy}, \textit{sad}, \textit{angry}, \textit{fearful}, \textit{disgusted}, and \textit{surprised}. In addition, each data sample is also annotated with a sentiment score on a Likert scale [-3, 3]. 

\paragraph{MUStARD.} The Multimodal Sarcasm Detection Dataset (MUStARD)~\citep{mustard} is a multimodal video corpus for sarcasm recognition. The dataset has 690 samples with an even number of sarcastic and non-sarcastic labels.

\vspace{2px}

We show the data statistics of three datasets in Table~\ref{tab:data-stats}.

\begin{table}[t!]
\centering
\resizebox{0.75\linewidth}{!}{
\begin{tabular}{l|ccc}
\toprule
Label & Train & Valid & Test \\
\midrule
\midrule
\multicolumn{4}{c}{\texttt{IEMOCAP}}\\\midrule
Neutral & 954 & 358 & 383 \\
Happy & 338 & 116 & 135 \\
Sad & 690 & 188 & 193 \\
Angry & 735 & 136 & 227 \\
\midrule
\multicolumn{4}{c}{\texttt{CMU-MOSEI}}\\\midrule
Angry & 3443 & 427 & 971 \\
Disgusted & 2720 & 352 & 922 \\
Fear & 1319 & 186 & 332 \\
Happy & 8147 & 1313 & 2522 \\
Sad & 3906 & 576 & 1334 \\
Surprised & 1562 & 201 & 479 \\
\midrule
\multicolumn{4}{c}{\texttt{MUStARD}}\\\midrule
Sarcastic & 276 & - & 69 \\
Non-sarcastic & 276 & - & 69 \\
\bottomrule\end{tabular}
}
\caption{Per-class data statistics of three datasets.}
\label{tab:data-stats}
\end{table}

\subsection{Data Feature Extraction}
Feature extraction is done for each modality to extract high-level features before training. 
For the textual modality, we use the pre-trained GloVe~\citep{pennington2014glove} embeddings to represent words (glove.840B.300d\footnote{\url{http://nlp.stanford.edu/data/glove.840B.300d.zip}}).
For the acoustic modality, COVAREP\footnote{\url{https://github.com/covarep/covarep}}~\citep{degottex2014covarep} is used to extract features of dimension 74 from the raw audio data. The features include fundamental frequency (F0), Voice/Unvoiced feature (VUV), quasi open quotient (QOQ), normalized amplitude quotient (NAQ), glottal source parameters (H1H2, Rd, Rd conf), maxima dispersion quotient (MDQ), parabolic spectral parameter (PSP), tilt/slope of wavelet response (peak/slope), harmonic model and phase distortion mean (HMPDM and HMPDD), and Mel Cepstral Coefficients (MCEP). 
For the visual modality, 35 facial action units~\citep{ekman1980facial} are extracted from each frame of the video with OpenFace 2.0\footnote{\url{https://github.com/TadasBaltrusaitis/OpenFace}}~\citep{baltrusaitis2018openface}. Following previous works~\citep{tsai2018learning,zadeh2018multimodal}, word-level alignment is done with P2FA~\citep{yuan2008p2fa} to achieve the same sequence length for each modality. We reduce multiple feature segments within one aligned word into a single segment by taking the mean value over the segments.

\subsection{Evaluation Metrics} \label{sec:metrics}
Overall, we use three metrics for different datasets. For the emotion recognition task, as we evaluate on each emotion category, there are much more negative data that positive data, we use the weighted Accuracy (WAcc)~\citep{wacc} to mitigate the class imbalance issue. The formula of WAcc is
\begin{align*}
    \text{WAcc} = \frac{TP \times N/P + TN}{2N}
\end{align*}
\noindent where $N$ means total negative, $TN$ true negative, $P$ total positive, and $TP$ true positive. For sentiment analysis and sarcasm recognition, we just use normal accuracy as the data is more balanced. In addition, we also use the F1-score for all the tasks. 

However, we do not use the binary weighted F1-score (WF1) as some previous works~\citep{mfn,zadeh2018multimodal,mult,rahman2020integrating} did. The formula of WF1 is shown below,
\begin{align*}
    \text{WF1} = \frac{P}{I} \times \text{F1}_p + \frac{N}{I} \times \text{F1}_n
\end{align*}
\noindent in which $\text{F1}_p$ is the F1 score that treats positive samples as \textit{positive}, while $\text{F1}_n$ treats negative samples as \textit{positive}, and they are weighted by their portion of the data ($I$ is the total number of samples).
We think that WF1 makes the class imbalance issue even severer, as $\text{F1}_p$ only contributes a small portion of the total WF1. According to our experiments on IMEOCAP,
when using WF1 to evaluate models, by increasing the threshold of classification, the WF1 increases a lot as well. Even when the threshold is 0.9 (means most of the samples will be classified as negative), the average of WF1 can still be higher than 0.7 while the WAcc is already below 0.5. A similar phenomenon is also observed by \citet{Dai2020ModalityTransferableEE}. Therefore, we just use the normal unweighted F1-score, which we think can better reflect the model's performance.

\section{Model Benchmarking} \label{sec:method_benchmark}
As mentioned in Section \ref{sec:intro}, we conjecture that the data sacacity issue of multimodal affect recognition tasks causes two problems: 1) models tend to be overfitted; and 2) instability of performance. To verify the severity of them, we benchmark 12 different models on two commonly used multimodal emotion recognition datasets: IEMOCAP~\citep{iemocap} and CMU-MOSEI~\citep{zadeh2018multimodal}. The models include six baselines, three recently proposed SOTA models, and three advanced baselines with the hybrid fusion method. All the models are listed in the first column of Table~\ref{tab:iemocap-result-full}. The implementation details are discussed in Section~\ref{sec:experiments}.

\begin{figure}[t!]
    \centering
    \includegraphics[width=\linewidth]{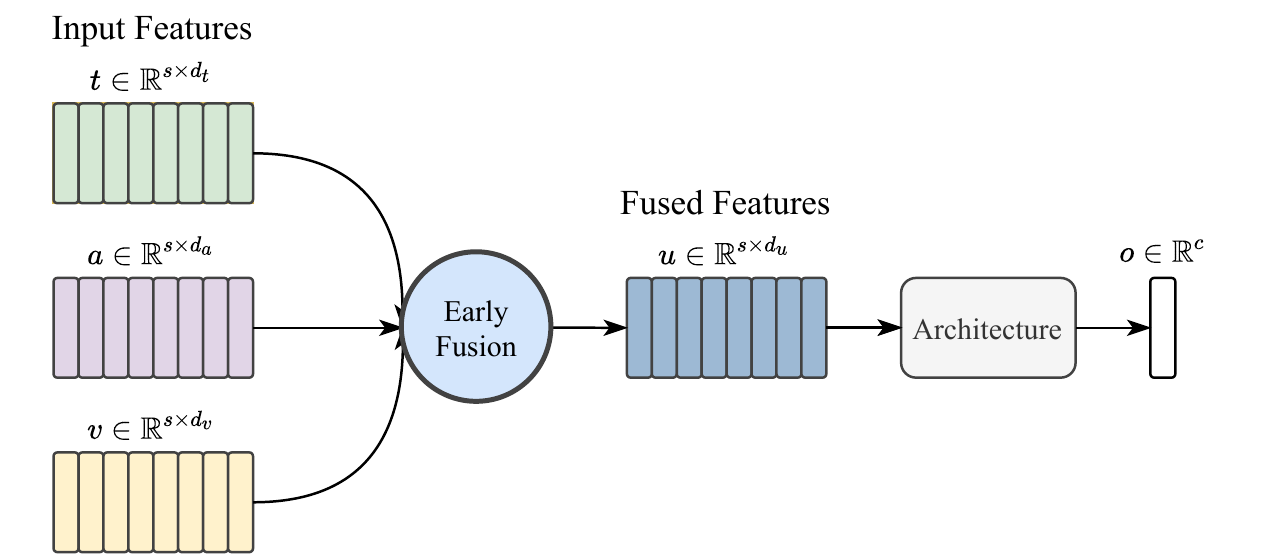}
    \small{\textbf{(a) Early fusion}}
    \vspace{5pt}

    \includegraphics[width=\linewidth]{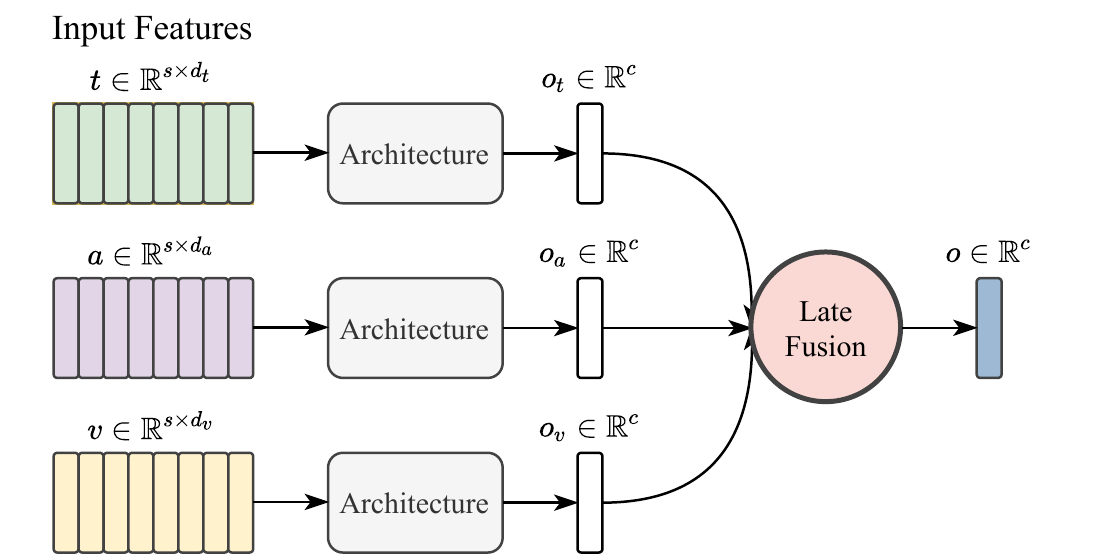}
    \vspace{-5pt}
    
    \small{\textbf{(b) Late fusion}}
    \vspace{10pt}
    
    \includegraphics[width=\linewidth]{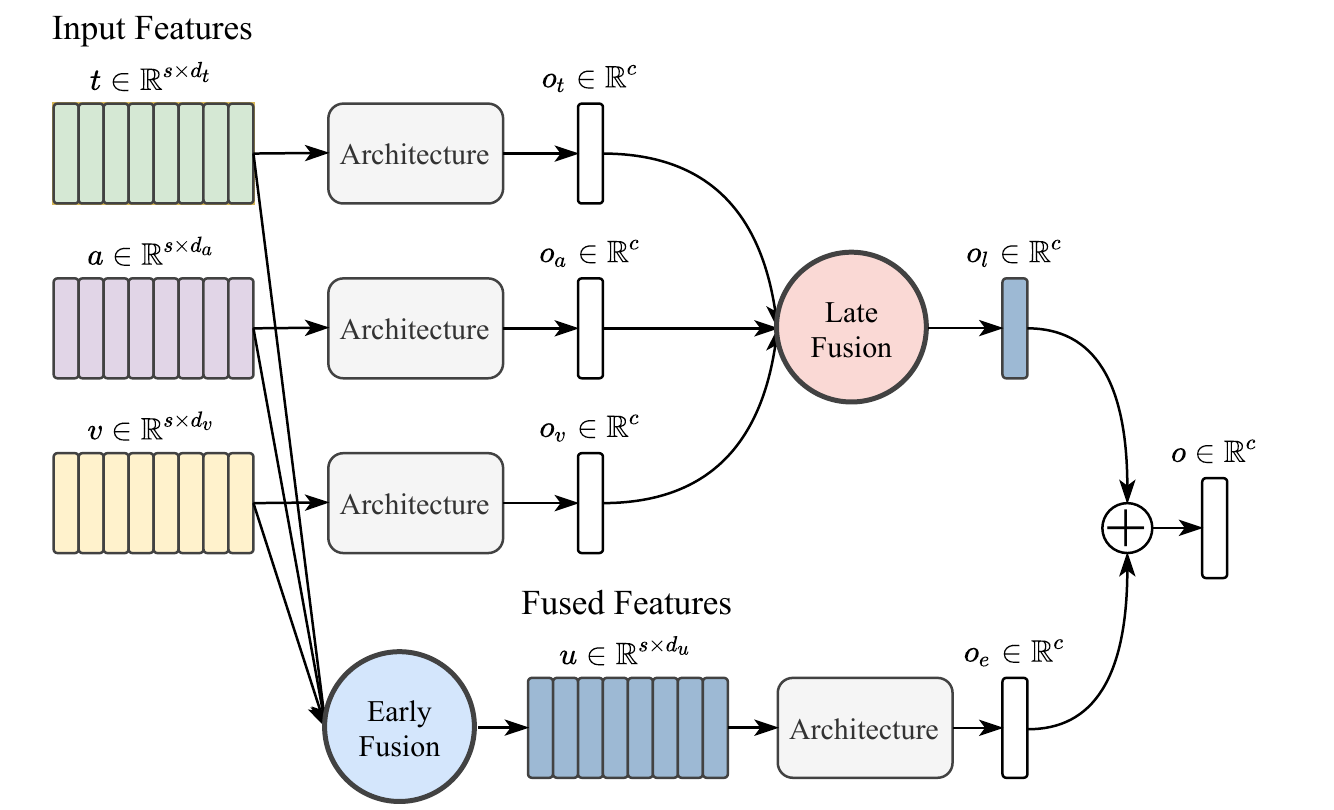}
    \vspace{-5pt}
    
    \small{\textbf{(c) Hybrid early-late fusion}}
    \vspace{0pt}
    
    \caption{Visualization of three model-agnostic modality fusion methods. (a) Early fusion: inputs from multiple modalities are fused at the entry level; (b) Late fusion: inputs from multiple modalities go through a dedicated model and then fused at the ouput level; and (c) Hybrid fusion: a combination of (a) and (b). Here, $\bigoplus$ represents the weighted sum operation.}
    \label{fig:baselines}
\end{figure}

\subsection{Baselines}
For the baselines, we use two commonly used model-agnostic fusion methods (Figure \ref{fig:baselines}): 1) early-fusion (EF), which can capture low-level cross-modal interactions; and 2) late-fusion (LF), which can capture high-level ones. They are used to build simple baselines. Besides, based on EF and LF, we propose to combine them as a hybrid fusion method (EF-LF) to construct strong baselines.

Within each fusion method, we apply three different model architectures, Average of features (AVG), bi-directional Long-Short Term Memory (LSTM)~\citep{lstm}, and Transformer~\citep{vaswani2017attention}, to process sequences of feature vectors. As the simplest baseline, the AVG model just takes the mean of the input vectors as the output vector. For LSTM, we take the output vector at the last time step as the representation of the whole input sequence. For Transformer, following the practice of previous work~\citep{vaswani2017attention,bert,roberta}, we prepend a \texttt{[CLS]} token to the input sequence and use the output embedding at that position. 

\subsection{SOTA Models} 
Apart from the baselines, we also select three recently proposed state-of-the-art models on this task for comparison: 1) Memory Fusion Network (MFN)~\citep{mfn}, which has a multi-view gated memory for storing cross-modal interactions over time; 2) Multimodal Transformer (MulT)~\citep{mult}, which fuses each pair of modalities with self-attention and can handle unaligned sequences; and 3) the model proposed by~\citet{Dai2020ModalityTransferableEE}, which leverages the information inside emotion embeddings (EMO-EMB) and can hanlde zero-/few-shot senarios.

\section{Weakly-supervised Multi-task Learning} \label{sec:mtl}
In this section, we first formally define the problem settings. 
Then, we explain our method for multi-task learning and how do we generate weak labels.

\subsection{Problem Definition}
We define a multimodal affect recognition dataset $D$ with $I$ data samples as $D=\{(t_i,a_i,v_i,y_i)\}^I_{i=1}$, in which $t_i \in \mathbb{R}^{s \times d_t}$ is a sequence of word embeddings to represent a sentence, $a_i \in \mathbb{R}^{s \times d_a}$ denotes an aligned sequence of audio features, $v_i \in \mathbb{R}^{s \times d_v}$ denotes an aligned sequence of facial action units (FAUs) extracted from the video frames, and $y_i \in \mathbb{R}^{d_y}$ denotes the golden label. Each modality has the same sequence $s$. For different datasets, the inputs $t, a, v$ have the same dimension as we use the same feature extraction across all datasets, only the dimension of labels $d_y$ might be different, depending on how many classes the dataset has.


\begin{table}[t!]
\centering
\resizebox{\linewidth}{!}{
\begin{tabular}{l|ccc|ccc}
\toprule
\multirow{2}{*}{\textbf{Dataset}} & \multicolumn{3}{c|}{\textbf{Strong Label}} & \multicolumn{3}{c}{\textbf{Weak Label}} \\
& Emo & Sen & Sar & Emo & Sen & Sar \\
\midrule
IEMOCAP & \checkmark & - & - & - & \small{MOSEI} & \small{MUStARD} \\
MOSEI & \checkmark & \checkmark & - & \small{IEMOCAP} & - & \small{MUStARD} \\
MUStARD & - & - & \checkmark & \small{MOSEI} & \small{MOSEI} & - \\
\bottomrule
\end{tabular}
}
\caption{
Emo, Sen, and Sar are abbreviations of Emotion, Sentiment, and Sarcasm, respectively. In the second column, \checkmark denotes the existence of human annotated strong labels, and - denotes the absence. In the third column, we show the existence of model generated weak labels for each dataset, and also which dataset the model is trained on.
}
\label{tab:dataset_labels}
\end{table}

\begin{table*}[t!]
\centering
\resizebox{\textwidth}{!}{
\begin{tabular}{l|cc|cc|cc|cc|cc}
\toprule
\multirow{2}{*}{\textbf{Model}} & \multicolumn{2}{c}{\textbf{Neutral}} & \multicolumn{2}{c}{\textbf{Happy}} & \multicolumn{2}{c}{\textbf{Sad}} & \multicolumn{2}{c}{\textbf{Angry}} & \multicolumn{2}{c}{\textbf{Average}} \\  
 & WAcc. (\%) & F1 (\%) & WAcc. (\%) & F1 (\%) & WAcc. (\%) & F1 (\%) & WAcc. (\%) & F1 (\%) & WAcc. (\%) & F1 (\%) \\
\midrule
\midrule
\multicolumn{11}{c}{\texttt{Simple Baselines}}\\\midrule
EF-AVG & 64.9 ± \small{0.8} & 67.9 ± \small{1.7} & 73.0 ± \small{3.2} & 79.8 ± \small{4.7} & 71.4 ± \small{2.4} & 58.6 ± \small{1.1} & 41.7 ± \small{3.7} & 50.3 ± \small{3.6} & 67.6 ± \small{4.7} & 54.6 ± \small{2.7} \\
EF-LSTM & 65.7 ± \small{4.6} & 66.0 ± \small{3.2} & 74.0 ± \small{1.8} & 81.9 ± \small{1.6} & 71.9 ± \small{2.4} & 64.1 ± \small{2.4} & 40.9 ± \small{4.7} & 54.8 ± \small{3.9} & 70.4 ± \small{1.9} & 57.6 ± \small{2.8} \\
EF-TRANS & 62.6 ± \small{4.0} & 66.9 ± \small{1.5} & 72.9 ± \small{2.3} & 77.8 ± \small{2.8} & 70.0 ± \small{2.2}  & 59.5 ± \small{3.5} & 39.5 ± \small{2.7} & 50.7 ± \small{2.3} & 64.3 ± \small{3.1} & 53.5 ± \small{2.2} \\
\midrule
LF-AVG  & 67.1 ± \small{0.5} & 66.8 ± \small{0.9} & 75.2 ± \small{1.0} & 81.4 ± \small{0.3} & 72.6 ± \small{0.4} & 58.3 ± \small{0.8} & 42.2 ± \small{1.3} & 53.2 ± \small{1.4} & 68.0 ± \small{0.8} & 55.4 ± \small{0.6} \\
LF-LSTM & 66.5 ± \small{1.5} & 68.0 ± \small{1.3} & 75.5 ± \small{0.8} & 78.5 ± \small{1.2} & 72.1 ± \small{0.7} & 63.2 ± \small{1.4} & 44.2 ± \small{1.5} & 56.1 ± \small{1.5} & 67.4 ± \small{1.3} & 57.7 ± \small{0.3} \\
LF-TRANS & 67.2 ± \small{0.5} & 65.8 ± \small{2.9} & 73.4 ± \small{1.6} & 78.3 ± \small{1.5} & 71.2 ± \small{1.3} & 62.4 ± \small{1.1} & 39.5 ± \small{2.8} & 53.0 ± \small{2.4} & 65.0 ± \small{3.0} & 55.0 ± \small{1.9} \\
\midrule
\multicolumn{11}{c}{\texttt{Strong Baselines (Hybrid fusion)}}\\\midrule
EF-LF AVG & 66.3 ± \small{1.1} & 68.4 ± \small{1.1} & 78.8 ± \small{0.8} & 82.5 ± \small{0.6} & 74.0 ± \small{0.3} & 55.4 ± \small{2.2} & \textbf{45.5 ± \small{1.6}} & 58.4 ± \small{1.0} & 67.8 ± \small{0.5} & 56.8 ± \small{0.6} \\
EF-LF LSTM  & \textbf{68.8 ± \small{0.9}} & \textbf{68.8 ± \small{1.2}} & 76.2 ± \small{2.0} & \textbf{83.8 ± \small{0.5}} & \textbf{74.4 ± \small{0.4}} & 64.2 ± \small{1.1} & 43.5 ± \small{1.9} & 57.9 ± \small{0.8} & \textbf{72.4 ± \small{2.1}} & 59.5 ± \small{0.2} \\
EF-LF TRANS & 68.4 ± \small{0.6} & 67.8 ± \small{0.7} & \textbf{79.0 ± \small{1.4}} & 81.8 ± \small{1.8} & 74.2 ± \small{0.8}  & 63.6 ± \small{0.6} & 41.2 ± \small{1.9} & 60.5 ± \small{2.5} & 68.8 ± \small{2.2} & 58.5 ± \small{1.3} \\
\midrule
\multicolumn{11}{c}{\texttt{SOTA Models}}\\\midrule
MULT & 65.4 ± \small{0.4} & 66.7 ± \small{1.7} & 75.5 ± \small{2.6} & 81.6 ± \small{1.4} & 72.3 ± \small{0.8} & 60.2 ± \small{0.3} & 43.4 ± \small{2.1} & 53.1 ± \small{2.9} & 70.1 ± \small{2.1} & 56.7 ± \small{1.0} \\
MFN & 68.3 ± \small{1.5} & 67.7 ± \small{1.0} & 74.1 ± \small{1.5} & 80.3 ± \small{0.7} & 72.6 ± \small{0.4} & \textbf{65.1 ± \small{1.4}} & 44.9 ± \small{1.3} & 55.4 ± \small{1.6} & 69.2 ± \small{2.0} & 58.7 ± \small{1.1} \\
EMO-EMB & 65.7 ± \small{1.3} & 68.8 ± \small{1.2} & 78.1 ± \small{2.5} & 82.8 ± \small{1.6} & 73.8 ± \small{0.5} & 61.8 ± \small{1.7} & 44.2 ± \small{2.9} & \textbf{61.5 ± \small{3.2}} & 71.4 ± \small{2.3} & \textbf{59.7 ± \small{1.6}} \\
\bottomrule
\end{tabular}
}
\caption{Performance evaluation of 12 models (six simple baselines, three strong baselines, and three SOTA models) on the IEMOCAP~\citep{iemocap} dataset. We report the weighted accuracy (WAcc) and the F1-score on four emotion categories: \textit{neutral}, \textit{happy}, \textit{sad}, and \textit{angry}. In addition, we also report the average of them as an overall measurement. For each model, we run five random seeds \{0, 1, 2, 3, 4\} and report the \texttt{mean} ± \texttt{standard\_deviation}. The best performance is decorated in bold.}
\label{tab:iemocap-result-full}
\end{table*}

\subsection{Multi-task Learning (MTL) for Multimodal Affect Recognition} \label{sec:mtlmar} 
As mentioned in Section~\ref{sec:intro}, datasets for multimodal affect recognition are relatively small in nature, which causes two problems. To mitigate them, we propose to utilize MTL by two reasons. Firstly, MTL can potentially improve the generalization performance when tasks are relevant~\citep{mtl_survey}. Secondly, there are many existing datasets related to multimodal affect recognition, even though each can be small, and their data come in a similar format (text, audio, video) and can be pre-processed in the same way.
Specifically, we leverage three relevant multimodal tasks: 1) emotion recognition; 2) sentiment analysis; and 3) sarcasm recognition. 

\paragraph{Weak Label Acquisition.} Given two separate datasets $D_1$ and $D_2$ on two different tasks $T_1$ and $T_2$, we generate weak labels of $T_2$ for $D_1$ in two steps: 1) first train a model on the data of $D_2$; and 2) use the trained model to infer predictions on the data of $D_1$, and the predictions are treated as weak labels of $T_2$. Specifically, in this paper, we train a dedicated EF-LF-LSTM model to generate weak labels on each of the three tasks following the aforementioned procedure. The details of what kinds of labels each dataset has after weak label acquisition are shown in Table~\ref{tab:dataset_labels}. For sentiment analysis, followin previous papers~\citep{zadeh2018multimodal,Liu2018EfficientLM,rahman2020integrating}, we simplify it to a two-class classification problem (either positive or negative) and generate binary labels. For the other tasks, we follow the original categories of the datasets. In addition, we also store the accuracy of the EF-LF-LSTM model on each dataset as a confidence score of the generated weak labels.



\paragraph{Weakly-supervised MTL.}
After getting the weak labels, we can conduct weakly-supervised learning for different tasks.
The training procedure is shown in Eq.\ref{eq:mtl_1} and \ref{eq:mtl_2}:
\begin{gather}
    r_i = f_{W_b}(t_i,a_i,v_i)  \label{eq:mtl_1} \\
    \min_{W_b, W_1, \dots, W_J} L = \sum^{I}_{i=1} \sum^{J}_{j=1} \lambda_j \cdot L_j(A_{W_{j}}(r_i), y_i) \label{eq:mtl_2}
\end{gather}
\noindent where $f_{W_b}$ is the backbone model with weights $W_b$ shared by all tasks, which generates a representation $r_i$ of the input data, and $J$ is the number of tasks. For each task $j$, there is a linear layer $A_{W_{j}}$ with parameters $W_{j}$ to perform affine transformation on $r_i$ to get the desired output dimension. The overall objective is to minimize the total loss $L$ of all tasks, and each task has a loss function $L_j$ with a weighting factor $\lambda_j$. This weighting factor can be either the confidence score mentioned in the last paragraph, or a hyper-parameter searched manually. For the strong labels of the main task, the confidence score is 1.

For datasets that contain multiple labels, we can directly apply this MTL procedure, while for unlabeled tasks, we perform MTL in a weakly-supervised way. For example, we can train a model on sentiment analysis and use it to predict sentiment scores for a sarcasm recognition dataset. Then, jointly use predicted weak labels with the original labels of the dataset for MTL.

\section{Experiments and Analysis} \label{sec:experiments}


\begin{table*}[t!]
\centering
\resizebox{0.9\linewidth}{!}{
\begin{tabular}{l|l|cc|cc|cc}
\toprule
\textbf{Target Task} & \textbf{Training Tasks} & \multicolumn{2}{c|}{\textbf{EF-LF AVG}} & \multicolumn{2}{c|}{\textbf{EF-LF LSTM}} & \multicolumn{2}{c}{\textbf{EF-LF TRANS}} \\
\toprule
& & \textbf{Avg.WAcc} & \textbf{Avg.F1} & \textbf{Avg.WAcc} & \textbf{Avg.F1} & \textbf{Avg.WAcc} & \textbf{Avg.F1} \\
\midrule
\multirow{4}{*}{\begin{tabular}[c]{@{}l@{}}Emotion\\ (IEMOCAP)\end{tabular}} & Emotion & 67.8 & 56.8 & 72.4 & 59.5 & 68.8 & 58.5 \\
& + Sentiment$_{(W)}$ & \textbf{69.1} & \textbf{58.4} & 73.3 & \textbf{60.4} & \textbf{69.8} & \textbf{59.3} \\
& + Sarcasm$_{(W)}$ & 68.6 & 57.3 & 72.8 & 59.9 & 69.1 & 58.6 \\
& All & 68.7 & 57.5 & \textbf{73.4} & 60.1 & 69.6 & \textbf{59.3} \\
\midrule\midrule
\multirow{4}{*}{\begin{tabular}[c]{@{}l@{}}Emotion\\ (CMU-MOSEI)\end{tabular}} & Emotion & 66.7 & 42.6 & 65.9 & 42.1 & 65.6 & 42.1 \\
& + Sentiment$_{(S)}$ & \textbf{67.2} & 43.2 & \textbf{66.5} & \textbf{42.7} & \textbf{66.2} & 42.3 \\
& + Sarcasm$_{(W)}$ & 66.9 & 43.3 & 66.1 & 42.4 & 66.1 & \textbf{42.4} \\
& All & 67.0 & \textbf{43.3} & 66.1 & \textbf{42.7} & 66.0 & 42.4 \\
\midrule\midrule
& & \textbf{Acc} & \textbf{F1} & \textbf{Acc} & \textbf{F1} & \textbf{Acc} & \textbf{F1} \\
\midrule
\multirow{4}{*}{\begin{tabular}[c]{@{}l@{}}Sarcasm\\ (MUStARD)\end{tabular}} & Sarcasm & 63.8 & 65.8 & 68.1 & 70.7 & 62.3 & 63.9 \\
& + Emotion$_{(W)}$ & 65.2 & 65.2 & \textbf{71.0} & \textbf{74.0} & \textbf{64.1} & 64.5 \\
& + Sentiment$_{(W)}$ & 65.2 & \textbf{67.6} & \textbf{71.0} & 73.3 & 63.0 & \textbf{65.8} \\
& All & \textbf{65.9} & 66.7 & 69.6 & 70.8 & 63.0 & 63.9 \\
\midrule\midrule
& & \textbf{Acc} & \textbf{F1} & \textbf{Acc} & \textbf{F1} & \textbf{Acc} & \textbf{F1} \\
\midrule
\multirow{6}{*}{\begin{tabular}[c]{@{}l@{}}Sentiment\\ (CMU-MOSEI)\end{tabular}} & Sentiment$^{1}$ & 70.9 & 69.7 & 70.8 & 69.1 & 70.6 & 69.5 \\
& + Emotion$_{(S)}$\(^2\) & 71.6 & \textbf{70.4} & 71.5 & 70.2 & 71.9 & \textbf{71.7} \\
& + Emotion$_{(W)}$\(^3\) & 71.7 & 70.1 & 71.8 & 70.6 & \textbf{72.3} & 70.9  \\
& + Sarcasm$_{(W)}$\(^4\) & 71.4 & 69.8 & 71.1 & 69.4 & 72.0 & 70.9 \\
& All$_{(1+2+4)}$ & 71.5 & 70.1 & 71.4 & \textbf{70.8} & 71.8 & 71.0 \\
& All$_{(1+3+4)}$ & \textbf{71.8} & \textbf{70.4} & \textbf{71.9} & 70.6 & \textbf{72.3} & 71.0 \\
\bottomrule
\end{tabular}
}
\caption{Experimental results of MTL. In the first column, we show the main target task and the corresponding dataset. In the second column, we show the tasks used in the training process. The symbol + means adding an auxiliary task in the training, besides the main target task. \textit{All} means using all the tasks above for training. $(S)$ or $(W)$ indicates whether this external supervision is strong or weak. For emotion recognition, \textit{Avg.} means the average of all emotion categories. In the bottom block, we also compare the effectiveness of using strong and weak label when conducting MTL.}
\label{tab:multitask-result-all}
\end{table*}

\subsection{Experimental Settings}
To ensure we make a fair comparison of the models, we perform an elaborate hyper-parameter search with the following strategies. Firstly, for each model, we try the combinations of four learning rates $\{1\mathrm{e}^{-3}, 5\mathrm{e}^{-4}, 1\mathrm{e}^{-4}, 5\mathrm{e}^{-5}\}$ and six batch sizes \{$16$, $32$, $64$, $128$, $256$, $512$\}, resulting in $24$ experiments. Among their hyper-parameter settings achiveing top-5 performance, we further try different model-dependent hyper-parameters, such as the hidden dimension, feed-forwad dimension, number of layers, dropout values, etc. For the previous state-of-the-art models, we also conduct a similar hyper-parameter search based on their reported numbers in the paper. To test the stability of models and eliminate the possible contingency caused by weights initialization, for the best setting of each model, we run five different random seeds \{0, 1, 2, 3, 4\} and report the mean and standard deviation. The best hyper-parameters are shown in Appendix B. For the emotion recognition, we use the binary cross-entropy loss as the data are multi-labeled (a person can have multiple), with a loss weight for positive samples to alleviate the data imbalance issue. For the sentiment prediction and sarcasm recognition, we use the cross-entropy loss. The Adam optimizer~\citep{adam} is used for all of our experiments with $\beta_1=0.9$, $\beta_2=0.999$ and a weight decay of $1\mathrm{e}^{-5}$. 
Our code is implemented in PyTorch~\citep{pytorch} and run on a single NVIDIA 1080Ti GPU.

\subsection{Benchmarking Results Analysis}
To investigate the two problems aforementioned in Section~\ref{sec:intro} and have an overall understanding of the performance of various models on multimodal affect recognition, we benchmark 12 models (Section~\ref{sec:method_benchmark}) on CMU-MOSEI and IEMOCAP. The results on IEMOCAP are shown in Table~\ref{tab:iemocap-result-full}, and the results on CMU-MOSEI are included in Appendix A. 
As explained in Section~\ref{sec:metrics}, we use slightly different evaluation metrics compared to previous works~\citep{mfn,mult} to better reflect the model performance.
First of all, we discover that SOTA models do not have an obvious advantage over the baselines on these two datasets. 
The hybrid modality fusion (EF+LF) with a simple architecture can surpass the SOTA models. We conjecture that the data scarcity issue makes complex architextures not able to show their full capacity. For example, the Multimodal Transformer (MULT)~\citep{mult} use a hidden dimension of only 40 and a dropout value of 0.3 to achieve its best performance (i.e. avoid overfitting).
Secondly, we find out that the performance is unstable given the small size of data. For example, by altering a random seed, the WAcc can change up to 4.7\%. On average, we get a standard deviation of 1.8\% for WAcc and 1.7\% for F1-score on the IEMOCAP.
Besides the dataset size, we speculate that another reason for the overfitting is that there is a feature extraction step before training to get the high-level features from the raw data. Thus the information in the resulting input features is highly concentrated, especially the audio and video features, which makes the problem of data scarcity even severer.

\subsection{Effects of Weakly-supervised MTL}


Experimental results of MTL are shown in Table~\ref{tab:multitask-result-all}. We evaluate weakly-supervised MTL using three models for four target tasks across three datasets. Generally, by incorporating auxiliary tasks with weak labels, we can achieve a better performance and this kind of improvement is consistantly observed on four target tasks across all datasets. We use the weak labels as soft labels by weighting the loss down from the auxiliary tasks. As mentioned in Section~\ref{sec:mtlmar}, the loss weights can be the confidence scores or manually searched. The best loss weights are reported in Appendix B. Additionally, it also mitigates the instability problem. Still on the emotion recognition task, the standard deviations of WAcc and F1-score are lowered by 0.4\% and 0.3\%, respectively.

\begin{table}[t]
\centering
\resizebox{\linewidth}{!}{
\begin{tabular}{l|ccccccc}
\toprule
\textbf{Label} & \textbf{Angry} & \textbf{Disgust} & \textbf{Fear} & \textbf{Happy} & \textbf{Sad} & \textbf{Surprised} & \textbf{Sarcasm} \\
\midrule
Sen & -0.39 & -0.56 & -0.01 & 0.53 & -0.34 & -0.07 & 0.05 \\
\bottomrule
\end{tabular}
}
\caption{Pearson correlation of sarcasm label and emotion label to the sentiment label on CMU-MOSEI dataset.}
\label{tab:correlation-sentiment}
\end{table}

On the multimodal emotion recognition task, we observe that either sentiment or sarcasm weak labels can help to increase the performance of the models. On the sarcasm recognition task, the improvement gained by MTL is even larger. For example, the EF-LF LSTM model can achieve 2.9\% accuracy and 3.3\% F1-score improvement when trained with an additional sentiment or emotion task. This fact shows that even though the auxiliary labels are noisy, MTL is still effective to help the model generalize better, especially when the target dataset has an insufficient number of samples. However, when we try to add two weak labels together to the main task, we do not observe further improvement compared to one auxiliary task. We spectulate that it is because more kinds of weak labels introduce more noise from different domains, which makes it harder for the model to learn and generalize.

On the multimodal sentiment analysis task, MTL with emotion labels consistently make a greater contribution than with sarcasm labels. 
This aligns with our label analysis in Table~\ref{tab:correlation-sentiment} that sentiment has a higher correlation with emotion than sarcasm.
Furthermore, we also compare the effectiveness of strong and weak emotion labels on the CMU-MOSEI dataset when sentiment analysis is the main task. We observe that weak emotion labels can contribute similarly or even slightly better than the strong emotion labels. One of the reason is that they contain different emotion categories, and the weak emotion labels have the \textit{neutral} class, which could be more helpful for recognizing sentiment. 

We think this is beneficial for future work, by providing the alternative to use weak-supervision as a cheaper solution for performance improvement compared to manually annotating the data.

\section{Conclusion and Future Work}

In this work, we prove the effectiveness of weakly-supervised multi-task learning (MTL) for multimodal affect recognition tasks. We show that it can significantly improve the performance, especially when the dataset size is small. For instance, on the sarcasm recognition task, the weakly-supervised MTL approach can improve the performance by up to 2.9\% accuracy and 3.3\% F1-score. We further conduct an empirical analysis on the effects of strong and weak (noisy) supervision, and show that weak supervision can help to boost performance almost as well as strong supervision. It's also a more flexible and cheaper way to incorporate more related supervision. Additionally, we introduce a simple but very effective hybrid modality fusion method by combining early fusion and late fusion. Its performance is on par with or even better than previous state-of-the-art models, which we believe can be used as a strong baseline for future work on multimodal affect recognition tasks.

\bibliography{custom}
\bibliographystyle{acl_natbib}

\appendix

\section{Benchmarking results on the CMU-MOSEI}
We show the benchmarking results on CMU-MOSEI in Table~\ref{tab:mosei-result-full}.

\begin{table*}[t!]
\centering
\resizebox{0.8\textwidth}{!}{
\begin{tabular}{l|ccccccc}
\toprule
\textbf{Model} & \textbf{Angry} & \textbf{Disgust} & \textbf{Fear} & \textbf{Happy} & \textbf{Sad} & \textbf{Surprised} & \textbf{Average} \\
\midrule
\multicolumn{8}{c}{Weighted Acc. (\%)} \\ 
\midrule
EF-AVG & 67.6 ± \small{0.3} & 73.2 ± \small{0.4} & 66.2 ± \small{0.8} & 68.1 ± \small{0.3} & 63.0 ± \small{0.4} & 61.1 ± \small{0.6} & 66.6 ± \small{0.2} \\
EF-LSTM & 66.3 ± \small{0.3} & 72.3 ± \small{0.2} & 67.1 ± \small{0.6} & 67.3 ± \small{0.4} & 63.0 ± \small{0.3} & 58.4 ± \small{0.3} & 65.8 ± \small{0.1} \\
EF-TRANS & 60.5 ± \small{0.8} & 71.1 ± \small{1.3} & 62.9 ± \small{1.1} & 65.6 ± \small{0.3} & 62.1 ± \small{0.6} & 55.7 ± \small{0.7} & 63.0 ± \small{0.4} \\
\midrule
LF-AVG  & 66.5 ± \small{0.6} & 72.8 ± \small{0.2} & 66.3 ± \small{0.8} & 68.3 ± \small{0.3} & 62.9 ± \small{0.5} & 61.6 ± \small{0.1} & 66.4 ± \small{0.3} \\
LF-LSTM & 66.3 ± \small{0.6} & 72.6 ± \small{0.3} & 66.7 ± \small{0.7} & 68.1 ± \small{0.1} & 62.5 ± \small{0.5} & 61.0 ± \small{0.6} & 66.2 ± \small{0.1} \\
LF-TRANS & 63.6 ± \small{1.0} & 72.1 ± \small{0.5} & 64.2 ± \small{1.2} & 67.2 ± \small{0.4} & 62.6 ± \small{0.5} & 60.9 ± \small{1.2} & 65.1 ± \small{0.4} \\
\midrule
MULT & 64.6 ± \small{0.7} & 73.4 ± \small{0.6} & 64.4 ± \small{1.2} & 68.9 ± \small{0.3} & 62.8 ± \small{0.2} & 61.2 ± \small{0.2} & 65.9 ± \small{0.4} \\
MFN & 66.2 ± \small{1.2} & 72.9 ± \small{0.4} & 65.5 ± \small{1.3} & 67.9 ± \small{0.4} & 62.1 ± \small{0.4} & 61.2 ± \small{0.8} & 66.0 ± \small{0.3} \\
EMO-EMB & 66.5 ± \small{0.1} & 73.3 ± \small{0.3} & 66.6 ± \small{1.3} & 67.8 ± \small{0.6} & 62.5 ± \small{0.5} & 61.2 ± \small{0.6} & 66.3 ± \small{0.4} \\
\midrule
EF-LF AVG & 66.9 ± \small{0.1} & 73.1 ± \small{0.1} & 66.8 ± \small{0.2} & 68.3 ± \small{0.5} & 62.9 ± \small{0.4} & 62.2 ± \small{0.4} & 66.7 ± \small{0.2} \\
EF-LF LSTM  & 66.3 ± \small{0.2} & 72.6 ± \small{0.6} & 66.2 ± \small{0.3} & 67.6 ± \small{0.4} & 63.0 ± \small{0.5} & 59.6 ± \small{0.5} & 65.9 ± \small{0.1} \\
EF-LF TRANS & 63.0 ± \small{0.4} & 72.4 ± \small{0.4} & 65.0 ± \small{0.2} & 68.6 ± \small{0.3} & 63.2 ± \small{0.6} & 61.4 ± \small{1.3} & 65.6 ± \small{0.1} \\
\midrule
\multicolumn{8}{c}{F1-Score (\%)} \\ 
\midrule
EF-AVG & 45.0 ± \small{0.4} & 49.9 ± \small{0.6} & 21.1 ± \small{0.6} & 64.5 ± \small{0.8} & 49.5 ± \small{0.3} & 23.6 ± \small{0.4} & 42.3 ± \small{0.2} \\
EF-LSTM & 43.9 ± \small{0.3} & 50.3 ± \small{0.2} & 21.8 ± \small{0.3} & 66.5 ± \small{0.7} & 48.7 ± \small{0.3} & 22.0 ± \small{0.2} & 42.2 ± \small{0.1} \\
EF-TRANS & 37.2 ± \small{1.4} & 50.1 ± \small{1.2} & 18.3 ± \small{0.6} & 65.2 ± \small{0.6} & 48.0 ± \small{0.5} & 20.0 ± \small{0.6} & 39.8 ± \small{0.4} \\
\midrule
LF-AVG  & 44.1 ± \small{0.6} & 49.6 ± \small{0.3} & 21.1 ± \small{0.8} & 67.1 ± \small{0.4} & 49.0 ± \small{0.6} & 24.3 ± \small{0.1} & 42.5 ± \small{0.4} \\
LF-LSTM & 44.0 ± \small{0.6} & 50.0 ± \small{0.4} & 22.1 ± \small{0.3} & 67.1 ± \small{0.4} & 48.3 ± \small{0.5} & 24.2 ± \small{0.5} & 42.6 ± \small{0.1} \\
LF-TRANS & 41.4 ± \small{1.1} & 49.8 ± \small{0.6} & 18.7 ± \small{0.4} & 67.9 ± \small{0.4} & 48.5 ± \small{0.4} & 23.9 ± \small{0.9} & 41.7 ± \small{0.3} \\
\midrule
MULT & 42.4 ± \small{0.8} & 50.5 ± \small{0.8} & 20.6 ± \small{0.6} & 68.9 ± \small{0.9} & 48.7 ± \small{0.2} & 23.8 ± \small{0.3} & 42.5 ± \small{0.3} \\
MFN & 43.8 ± \small{1.0} & 50.1 ± \small{0.6} & 21.1 ± \small{1.5} & 67.0 ± \small{1.5} & 48.1 ± \small{0.5} & 24.4 ± \small{0.7} & 42.4 ± \small{0.3} \\
EMO-EMB & 44.0 ± \small{0.2} & 50.2 ± \small{0.4} & 21.2 ± \small{0.4} & 66.3 ± \small{0.8} & 48.6 ± \small{0.8} & 23.9 ± \small{0.3} & 42.4 ± \small{0.3} \\
\midrule
EF-LF AVG & 44.4 ± \small{0.2} & 49.9 ± \small{0.1} & 21.1 ± \small{0.4} & 66.4 ± \small{1.2} & 49.2 ± \small{0.4} & 24.3 ± \small{0.4} & 42.6 ± \small{0.3} \\
EF-LF LSTM  & 43.9 ± \small{0.2} & 50.2 ± \small{0.7} & 20.9 ± \small{0.6} & 66.2 ± \small{0.9} & 48.8 ± \small{0.5} & 22.8 ± \small{0.4} & 42.1 ± \small{0.1} \\
EF-LF TRANS & 40.8 ± \small{0.6} & 50.3 ± \small{0.5} & 19.0 ± \small{0.3} & 69.2 ± \small{0.4} & 48.7 ± \small{0.8} & 24.6 ± \small{1.0} & 42.1 ± \small{0.1} \\
\bottomrule
\end{tabular}
}
\caption{Performance evaluation of different models on the CMU-MOSEI dataset. We report the weighted accuracy (WAcc) and the F1-score on six emotion categories: \textit{angry}, \textit{disgust}, \textit{fear}, \textit{happy}, \textit{sad} and \textit{surprised}.}
\label{tab:mosei-result-full}
\end{table*}

\section{Best Hyper-parameters}
We include the best hyper-parameters from our hyper-parameter search in Table~\ref{tab:best_hp_iemocap} and Table~\ref{tab:best_hp_mosei}. In addition, we also show the best loss weights we used for MTL in Table~\ref{tab:multitask-loss-weights}.

\begin{table*}[]
\centering
\resizebox{0.8\textwidth}{!}{
\begin{tabular}{l|c|c|c|l}
\toprule
Model & LR & BS & Seed & Model Specific Hyper-params \\ \midrule
EF\_AVG & 1e-3 & 64 & 1 & hidden\_dim=128 \\
LF\_AVG & 1e-3 & 32 & 3 & hidden\_dims={[}128,32,16{]} \\
EF\_LSTM & 5e-4 & 128 & 0 & lstm\_dim=300, bidirect, layer=1, dropout=0.1 \\
LF\_LSTM & 1e-3 & 128 & 0 & lstm\_dim={[}300,128,128{]}, bidirect, layer=1, dropout=0.1 \\
EF\_TRANS & 5e-5 & 64 & 0 & layers=2, heads=5, ff\_dim=512, dropout=0.1 \\
LF\_TRANS & 1e-4 & 128 & 0 & layers=2, heads={[}4,2,2{]}, ff\_dim={[}300,128,64{]}, dropout=0.1 \\
MFN & 1e-3 & 128 & 0 & Refer to the original paper~\citep{mfn} \\
MULT & 2e-4 & 32 & 1 & Refer to the original paper~\citep{mult} \\
EMO\_EMB & 5e-4 & 64 & 0 & Refer to the original paper~\citep{Dai2020ModalityTransferableEE} \\
EF\_LF\_AVG & 5e-4 & 64 & 0 & - \\
EF\_LF\_LSTM & 5e-4 & 128 & 0 & - \\
EF\_LF\_TRANS & 1e-4 & 256 & 1 & - \\ \bottomrule
\end{tabular}
}
\caption{Best hyper-parameters of each model on the IEMOCAP dataset. LR, BS, represents the learning rate and batch size, respectively. For the hybrid fusion (EF+LF), we just use hyper-parameters from their corresponding EF and LF settings.}
\label{tab:best_hp_iemocap}
\end{table*}

\begin{table*}[]
\centering
\resizebox{0.8\textwidth}{!}{
\begin{tabular}{l|l|l|l|l}
\toprule
Model & LR & BS & Seed & Model Specific Hyper-params \\ \midrule
EF\_AVG & 5e-4 & 256 & 1 & hidden\_dim=128 \\
LF\_AVG & 5e-4 & 256 & 0 & hidden\_dims={[}128,64,32{]} \\
EF\_LSTM & 5e-5 & 32 & 2 & lstm\_dim=512, bidirect, layer=1, dropout=0.1 \\
LF\_LSTM & 5e-5 & 32 & 1 & lstm\_dim={[}300,128,128{]}, bidirect, layer=1, dropout=0.1 \\
EF\_TRANS & 5e-5 & 32 & 0 & layers=2, heads=5, ff\_dim=512, dropout=0.1 \\
LF\_TRANS & 5e-5 & 64 & 0 & layers=2, heads={[}4,2,2{]}, ff\_dim={[}300,128,64{]}, dropout=0.1 \\
MFN & 1e-4 & 128 & 0 & Refer to the original paper~\citep{mfn} \\
MULT & 1e-4 & 32 & 1 & Refer to the original paper~\citep{mult} \\
EMO\_EMB & 5e-5 & 256 & 0 & Refer to the original paper~\citep{Dai2020ModalityTransferableEE} \\
EF\_LF\_AVG & 5e-4 & 256 & 0 & - \\
EF\_LF\_LSTM & 1e-4 & 32 & 0 & - \\
EF\_LF\_TRANS & 5e-5 & 64 & 1 & - \\ \bottomrule
\end{tabular}
}
\caption{Best hyper-parameters of each model on the CMU-MOSEI dataset. LR, BS, represents the learning rate and batch size, respectively. For the hybrid fusion (EF+LF), we just use hyper-parameters from their corresponding EF and LF settings.}
\label{tab:best_hp_mosei}
\end{table*}

\begin{table*}[t!]
\centering
\resizebox{0.7\textwidth}{!}{
\begin{tabular}{l|l|c|c|c}
\toprule
\textbf{Target} & \textbf{Train} & \textbf{EF-LF AVG} & \textbf{EF-LF LSTM} & \textbf{EF-LF TRANS} \\
\bottomrule
\toprule
& & \textbf{Loss weights} & \textbf{Loss weights} & \textbf{Loss weights} \\
\midrule
\midrule
\multirow{4}{*}{\begin{tabular}[c]{@{}l@{}}Emotion\\ (IEMOCAP)\end{tabular}} & Emotion & - & -  & - \\
& +Sentiment$_{(W)}$ & 1.0 0.6 0.0 & 1.0 0.8 0.0 & 1.0 0.7 0.0   \\
& +Sarcasm$_{(W)}$ & 1.0 0.0 0.5 & 1.0 0.0 0.6 & 1.0 0.0 0.7  \\
& All & 1.0 0.8 0.3 &  1.0 0.7 0.5 & 1.0 0.7 0.3 \\
\midrule\midrule
\multirow{4}{*}{\begin{tabular}[c]{@{}l@{}}Emotion\\ (CMU-MOSEI)\end{tabular}} & Emotion & - & -  & - \\
& +Sentiment$_{(S)}$ & 1.0 0.4 0.0 & 1.0 1.0 0.0 & 1.0 1.0 0.0   \\
& +Sarcasm$_{(W)}$ & 1.0 0.0 0.6 & 1.0 0.0 0.8 & 1.0 0.0 0.5  \\
& All & 1.0 0.7 0.5 &  1.0 0.8 0.6 & 1.0 0.9 0.4 \\
\midrule\midrule
\multirow{4}{*}{\begin{tabular}[c]{@{}l@{}}Sarcasm\\ (MUStARD)\end{tabular}} & Sarcasm & -  & -  & -  \\
& +Emotion$_{(W)}$ & 0.4 0.0 1.0 & 0.4 0.0 1.0 & 1.0 0.0 1.0 \\
& +Sentiment$_{(W)}$ & 0.0 1.0 1.0 & 0.0 0.5 1.0 & 0.0 0.6 1.0  \\
& All & 0.4 1.0 1.0 & 0.1 0.5 1.0 & 1.0 0.1 1.0 \\
\midrule\midrule
\multirow{4}{*}{\begin{tabular}[c]{@{}l@{}}Sentiment\\ (CMU-MOSEI)\end{tabular}} & Sentiment\(^1\) & - & - & - \\
& +Emotion$_{(S)}$\(^2\) & 0.8 1.0 0.0 & 1.0 1.0 0.0 & 0.8 1.0 0.0 \\
& +Emotion$_{(W)}$\(^3\) & 0.6 1.0 0.0 & 0.6 1.0 0.0 & 0.8 1.0 0.0 \\
& +Sarcasm$_{(W)}$\(^4\) & 0.0 1.0 0.6 & 0.0 1.0 0.6 & 0.0 1.0 0.2 \\
& All$_{(1+2+4)}$ & 0.8 1.0 0.8 & 0.1 1.0 0.1 & 0.8 1.0 0.2 \\
& All$_{(1+3+4)}$ & 1.0 1.0 1.0 & 1.0 1.0 0.5 & 1.0 1.0 0.5 \\
\bottomrule
\end{tabular}
}
\caption{The loss weights $\lambda$ we used for each multi-tasking setting. Each entry has 3 numbers, which are the loss weights for the multimodal emotion recognition, sentiment analysis, sarcasm recognition, respectively. Therefore, $0.0$ means that task is not used in that experiment.}
\label{tab:multitask-loss-weights}
\end{table*}

\end{document}